\documentclass{article}
\usepackage{amsmath,epsfig}
\usepackage{times}
\usepackage{multirow}
\usepackage{subfigure}
\usepackage{helvet}
\usepackage{courier}

\usepackage{epsfig}
\usepackage{graphicx}
\usepackage{amsmath}
\usepackage{amssymb}
\usepackage{eso-pic}  
\usepackage{helvet}  
\usepackage{courier}  
\usepackage{url}  
\usepackage{graphicx}  
\usepackage{mathrsfs}
\usepackage{multirow}
\usepackage{subfigure}
\usepackage{comment}
\usepackage{bbm}
\usepackage{enumerate}

\usepackage{xcolor}
\usepackage{soul}
\usepackage[utf8]{inputenc}
\usepackage[small]{caption}
\usepackage{marvosym}
\usepackage{booktabs}       
\usepackage{amsfonts}       
\usepackage{nicefrac}       
\usepackage{microtype}      

\usepackage[pagebackref=true,breaklinks=true,letterpaper=true,colorlinks,bookmarks=false]{hyperref}

\usepackage[preprint]{spconfa4}

\copyrightnotice{Copyright notice 978-1-7281-1331-9/20/\$31.00 2020 IEEE}

\let\OLDthebibliography\thebibliography
\renewcommand\thebibliography[1]{
  \OLDthebibliography{#1}
  \setlength{\parskip}{0pt}
  \setlength{\itemsep}{0pt plus 0.3ex}
}

\begin{document}\sloppy

\def\x{{\mathbf x}}
\def\L{{\cal L}}

\title{Multi-Task Learning of Generalizable Representations for \\Video Action Recognition}
%
\name{Zhiyu Yao$^{*}$\thanks{* Equal contribution}, Yunbo Wang$^*$, Mingsheng Long (\Letter), Jianmin Wang, Philip S. Yu, and Jiaguang Sun}
\address{School of Software, BNRist, Tsinghua University, China \\
Research Center for Big Data, Tsinghua University, China\\
Beijing Key Laboratory for Industrial Big Data System and Application\\
{\tt\small \{yaozy19,wangyb15\}@mails.tsinghua.edu.cn, \{mingsheng,jimwang,psyu,sunjg\}@tsinghua.edu.cn} 
}

\maketitle

\begin{abstract}
In classic video action recognition, labels may not contain enough information about the diverse video appearance and dynamics, thus, existing models that are trained under the standard supervised learning paradigm may extract less generalizable features. We evaluate these models under a cross-dataset experiment setting, as the above \emph{label bias} problem in video analysis is even more prominent across different data sources. We find that using the optical flows as model inputs harms the generalization ability of most video recognition models.

Based on these findings, we present a multi-task learning paradigm for video classification. Our key idea is to avoid label bias and improve the generalization ability by taking data as its own supervision or supervising constraints on the data. First, we take the optical flows and the RGB frames by taking them as auxiliary supervisions, and thus naming our model as Reversed Two-Stream Networks (\textbf{Rev2Net}). Further, we collaborate the auxiliary flow prediction task and the frame reconstruction task by introducing a new training objective to Rev2Net, named Decoding Discrepancy Penalty (\textbf{DDP}), which constraints the discrepancy of the multi-task features in a self-supervised manner. Rev2Net is shown to be effective on the classic action recognition task. It specifically shows a strong generalization ability in the cross-dataset experiments.


\end{abstract}
\vspace{-5pt}
\section{Introduction}
\vspace{-5pt}

Learning generalizable representations is a new direction and yet an important problem in video analysis, considering that a good action recognition model should be able to handle various video environments. Different from most previous methods for video data, this paper discusses more about the generalization ability. 
Because of the \emph{label bias} problem that coarse video-level labels may only express short snippets of the entire untrimmed videos (within a single video), and may not contain enough information for the various video scenarios, including frame appearance and long-term action dynamics (across different videos). Thus, the traditional strongly supervised learning paradigm that is used by most existing video action models \cite{Simonyan14,wang2016temporal,carreira2017quo} may suffer from the label bias problem and leads to less generalizable spatiotemporal features.

\begin{figure}[t]
\centering
 \subfigure[Two-Stream 2D/3D CNNs]{
 \centering
    \includegraphics[height=0.37\columnwidth]{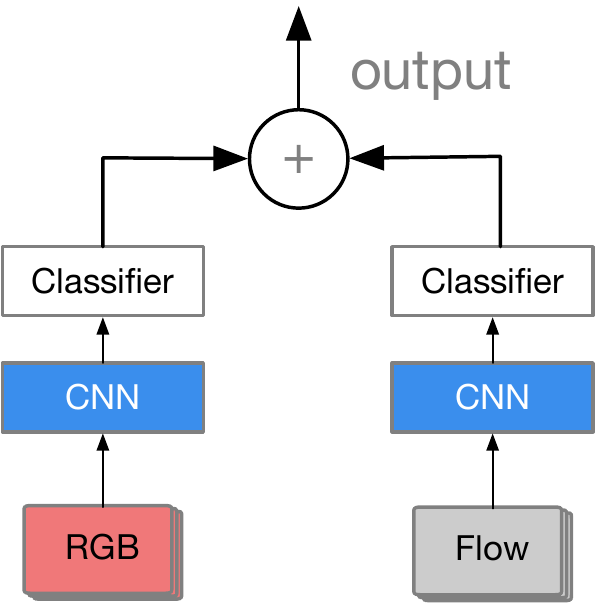}
    \label{1a}
    }
   \subfigure[Rev2Net]{
  \centering
    \includegraphics[height=0.37\columnwidth]{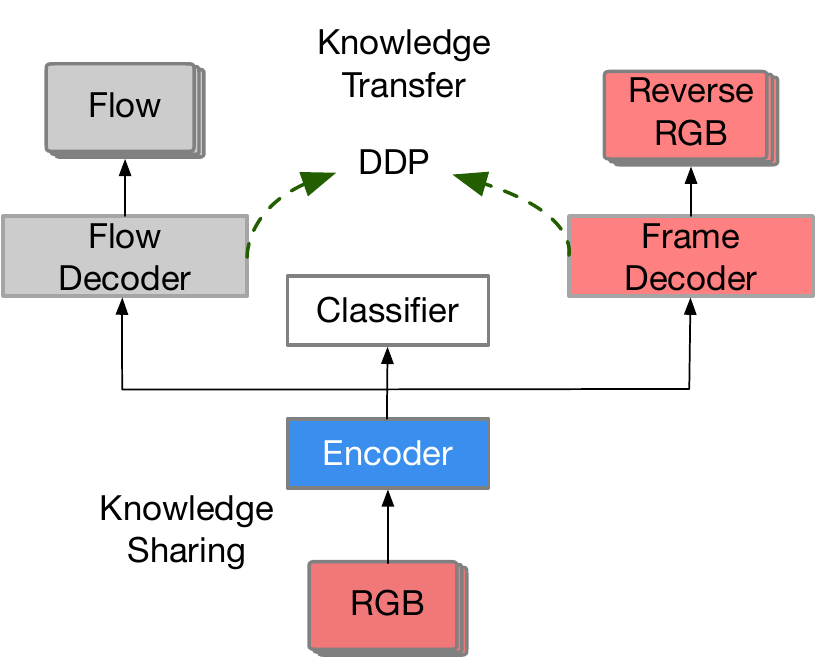}
    \label{1d}
  }
  \vspace{-5pt}
\caption{We propose a self-supervised multi-task learning framework to learn more generalizable video representations for action recognition. The proposed model, Rev2Net, differs from previous two-stream networks from the perspectives of knowledge transfer and knowledge sharing in the training procedure.}
\label{F3D*}
\vspace{-8pt}
\end{figure}

In this paper, we first evaluate the generalization ability of the most widely used video classification models \cite{wang2016temporal,carreira2017quo,lin2018temporal} in a cross-dataset setting, where it exists more severe label bias problem caused by different data sources. We find that these two-stream networks that are based on multi-modality inputs degenerate in such experiment settings. Particularly, we observe that the reason of this issue is that the optical flow features are \textbf{NOT generalizable}. Therefore, the multi-modality video data is not used properly from the perspective of domain generalization, and so solving this problem requires a different learning paradigm.

Based on the above intuitive and empirical motivations, we present a new self-supervised, multi-task learning approach for video action recognition. The key idea of this approach is to improve the generalization ability of the learned features by leveraging the multi-modality data as its own supervisions, and supervising the specific constraints on the data. Concretely, we propose a new multi-task network architecture named Reversed Two-Stream Networks (\textbf{Rev2Net}). As shown in Figure \ref{1d}, we train this network to classify video sequence, predict optical flows from RGB frames, and reconstruct the frames in a reverse order simultaneously. Rev2Net enables knowledge transfer from the auxiliary self-supervised tasks to the core supervised task through a common frame encoder. To further enhance the knowledge sharing between flow prediction and frame reconstruction task, we attempt to constrain the discrepancy of the two flow/frame decoders by applying a new training objective named Decoding Discrepancy Penalty (\textbf{DDP}) to both high-level and low-level features of Rev2Net. Both the ideas of knowledge sharing and knowledge transfer are shown to benefit the generalization ability. Rev2Net achieves competitive results on classic action recognition tasks within each of the UCF-101, HMDB-51, and Kinetics datasets. It also achieves the best performance in our cross-dataset experiment.

To sum up, this paper has the following contributions: 
\begin{itemize}
\vspace{-5pt}
  \item We design a cross-dataset experiment to evaluate the generalization ability of video recognition models, and observe that the features extracted from the optical flow data are less generalizable.
  \vspace{-5pt} 
  \item We present a new model named Rev2Net that can learn more generalizable features through a new multi-task learning framework with self-supervisions.
  \vspace{-5pt}
  \item We propose the DDP loss to encourage the collaboration of multiple auxiliary self-supervised tasks.
\end{itemize}

\vspace{-5pt}
\section{Related Work}
\vspace{-5pt}

Ever since the significant impact of CNNs upon image classification, many researchers have been trying out various CNN architectures for video action recognition, including 2D CNNs, 3D CNNs, and non-local modules \cite{NonLocal2018}. For all these models, the two-stream network framework with multi-modality inputs has been most widely explored and proved to be effective.
\vspace{-5pt}
\vspace{10pt} \noindent \textbf{Two-Stream CNN Models.} 
The two-stream networks were first introduced to video analysis by Simonyan \textit{et al.} \cite{Simonyan14}, which averages the classification results of a sub-network with frame inputs and another sub-network with pre-computed optical flow inputs.
The optical flow sub-network brings in significant performance gains due to the capability of capturing short-term video dynamics.
Since then, two-stream networks have been widely employed by many action recognition models \cite{Feichtenhofer16,wang2016temporal,wang2017spatiotemporal,wang2017appearance}, including 3D CNN models.

\vspace{-5pt}
\vspace{10pt} \noindent \textbf{3D CNN Models.} More recently, 3D CNNs have been widely explored \cite{Ji13,Tran15,carreira2017quo,qiu2017learning,xie2018rethinking}. These models realized unified modeling of the spatiotemporal features. But the 3D-Convs bring an inevitable increase in the number of network parameters, making these models hard to train. Carreira \textit{et al.} proposed the I3D model \cite{carreira2017quo} that inflates the filters of the 2D-Convs into 3D, making this model implicitly pretrained on ImageNet. Note that our proposed Rev2Net is mainly based on I3D. In Section \ref{sec:cross}, we will show that the above two-stream CNNs and 3D CNNs learn less generalizable features in a cross-dataset experiment. This paper attempts to solve this problem.

\vspace{-5pt}
\vspace{10pt} \noindent \textbf{Optical Flow Prediction and Multi-task Models.} 
Using optical flows as model inputs has been proved to be crucial for the performance of two-stream networks. However, some recent methods \cite{sevilla2017integration,Sun_2018_CVPR,ng2018actionflownet} showed that training to generate optical flows with some deep networks, e.g., FlowNet \cite{dosovitskiy2015flownet} and SpyNet \cite{ranjan2017optical}, could further improve the recognition performance.
In this paper, we go beyond optical flow estimation and propose a new method to collaborate multiple auxiliary tasks.

\vspace{-5pt}
\section{Analysis of Generalization Ability}
\label{sec:cross}
\vspace{-5pt}

We evaluate the generalization ability of video action recognition models with the cross-dataset experiments, for the reason that in such settings, the \emph{label bias} problem is more prominent. The videos from different datasets may have distinct scenes and action dynamics. Robust action recognition models should be easily adapted from one dataset to another.

\begin{figure}[h]
\begin{center}
\includegraphics[width=0.9\columnwidth]{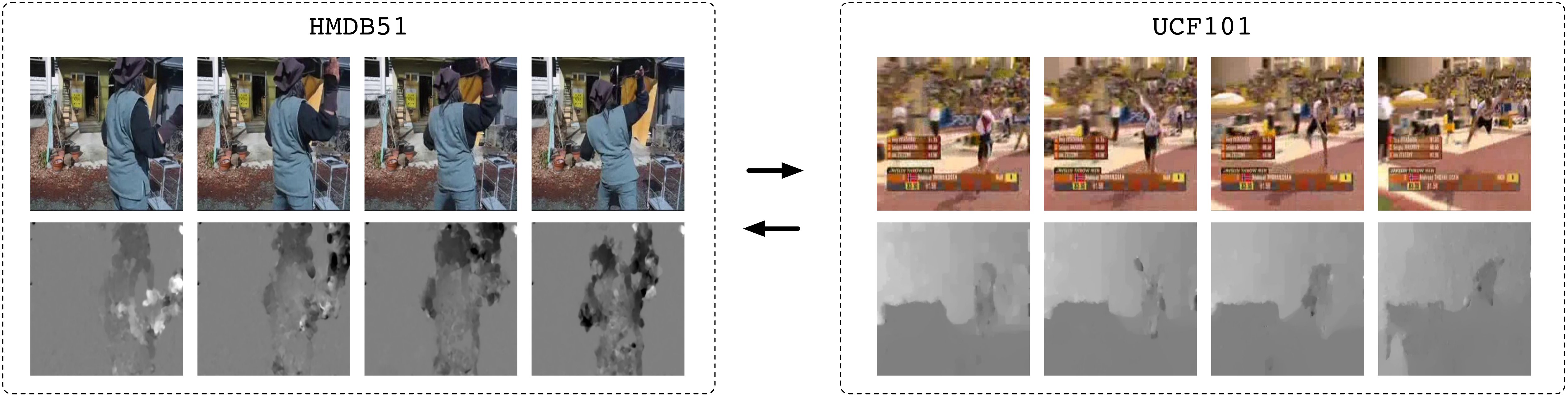}
\end{center}
\vspace{-10pt}
 \caption{Videos under the same category (``Archery'' in this case) from different datasets have diverse frame scenes and action dynamics, which can better evaluate the generalization ability of the action recognition models. Good video representations should be generalizable to the cross-dataset variations.}
  \label{transfer-sample}
  \vspace{-5pt}
\end{figure}

\vspace{-5pt}
\vspace{10pt} \noindent \textbf{Cross-Dataset Experiment.}
We design the cross-dataset experiment to verify the generalization ability of video action recognition models, adapting the domain adaptation settings from image data to video data. In general, we need to learn a model from the source dataset and to apply this model on the target dataset. More concretely, we select $16$ related categories from the UCF101 \cite{soomro2012ucf101} and HMDB51 \cite{Kuehne12} datasets. Note that these two datasets have diverse data patterns: HMDB51 is mostly collected from movies, while UCF101 is collected from YouTube and appears to be closer to real life. Even for the same action category, the video appearances of these two datasets are quite different, e.g. the scene complexity and the camera angles as shown in Figure \ref{transfer-sample}. Under these circumstances, it would be more likely to learn invariant features across datasets from similar motion cues.

We include the most prevalent action recognition networks in our cross-dataset experiment: TSN \cite{wang2016temporal}, TSM \cite{lin2018temporal} for 2D CNNs, and I3D \cite{carreira2017quo} for 3D CNNs. To further boost the generalization ability of these models, we incorporate a domain adaptation method that was originally proposed for image data \cite{ganin2015unsupervised}, which closes the distributions by matching the mean embeddings in the feature space across different domains.

Table \ref{transfer} shows the cross-dataset recognition results of the evaluated models with one or two modalities of input data. Surprisingly, we observe that for both 2D CNNs and 3D CNNs, the frame network consistently outperforms the flow network, which is against our experience on the classic video action recognition.
Consequently, the overall recognition accuracy of the two-stream networks yields no further improvement compared with the one-stream RGB network. This observation violates our expectations and our perceptions about the two-stream architecture on the classic video classification task.
Under the framework of the traditional two-stream networks that take optical flow as inputs, the only way to improve the overall cross-dataset performance is to improve the generalization ability of the optical flow stream. 
From these results, we may conclude that neither 2D CNNs nor 3D CNNs shows great generalization ability with optical flow inputs. 
We conjecture that optical flow data conveys very different action dynamics across datasets, leading to less generalizable features. Later, we will show that our proposed Rev2Net can improve the performance in the same cross-dataset experiment.

\begin{table}[t]
  \centering
  \caption{We explore the generalization ability of existing action recognition models using the cross-dataset experiment. We specifically apply a domain adaptation method DANN \cite{ganin2015unsupervised} to further enhance the generalization ability of these models.}
  \vspace{5pt}
  \label{transfer}
  \resizebox{\columnwidth}{!}{
  \begin{tabular}{cccc}
    \toprule
    Model & Input & HMDB $\rightarrow$ UCF & UCF $\rightarrow$ HMDB \\
    \midrule
    TSN & RGB         & 58.5 & 40.1  \\
    TSN & Flow      & 33.2 & 23.0  \\
    TSN & RGB + Flow & 60.9 & 40.3  \\
    \midrule
    TSM & RGB   & 58.9&  41.2\\
    TSM & Flow  & 35.1 &  24.0 \\
    TSM & RGB + Flow & 61.2 &  41.5 \\
    \midrule
    I3D & RGB         & 56.4 & 41.0  \\
    I3D & Flow       & 45.0 & 31.1  \\
    I3D & RGB + Flow  & 57.9 & 41.5  \\ 
    \bottomrule
  \end{tabular}
  }
\end{table}

\vspace{-5pt}
\section{Reversed Two-Stream Networks}
\vspace{-5pt}

In this section, we first present the Reversed Two-Stream Networks (Rev2Net) trained in a multi-task framework with self-supervisions. We then discuss a new training objective to further constrain the discrepancy and encourage the collaboration between the multiple auxiliary tasks of Rev2Net.

\vspace{-10pt}
\begin{figure*}[t]
  \centering
  \includegraphics[width=1\textwidth ]{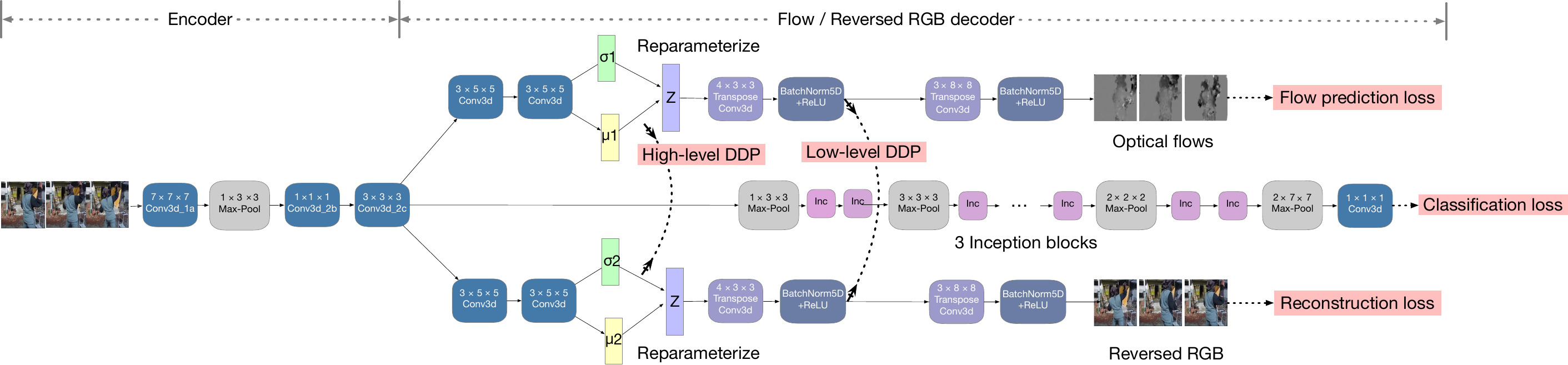}
  \caption{A schematic of the Rev2Net with decoding discrepancy penalty (DDP). We adopt the inception block from the inflated Inception-V1 \cite{carreira2017quo}. 
  The two decoder streams predicts corresponding optical flow and reconstruct the RGB inputs in a reversed order. 
  DDP is used for overcoming the discrepancy of two decoder streams, allowing them to be trained collaboratively.}
  \label{F3D}
  \vspace{-5pt}
\end{figure*}

\vspace{-5pt}
\subsection{A Self-Supervised Multi-Task Learning Framework} 
\vspace{-5pt}

As the optical flow stream shows limited generalization ability in cross-dataset setting, this stream severely affects the overall performance of the two-stream model. Thus, using optical flow as inputs is not an ideal approach for learning generalizable video features.
In contrast, our approach distills more generalizable knowledge by taking data as its own supervision or supervising constraints on the data. 
Specifically, we present the Reversed Two-Stream Networks (Rev2Net), which is trained in a multi-task learning framework with self-supervision from the multi-modality data.
The schematic of Rev2Net is shown in Figure \ref{F3D}. Rev2Net has four components: one encoder stream that only takes RGB frames as inputs, one classifier for action recognition, one decoder stream for optical flow prediction, and another decoder stream for reversed frame reconstruction.
The encoder stream operates on consecutive $32$ video frames. Along with the classifier, it has the same architecture as I3D. 
The two decoder streams are composed of 3D transpose convolutions. The flow decoder aims to emphasize learning the \textbf{short-term motion} features as well the foreground appearance features by using corresponding flow fields as supervisions. The frame decoder, on the other hand, aims to emphasize learning the \textbf{long-term motion} features by reconstructing the input frames in a reversed order from encoded 3D feature volume, which can be viewed as an information bottleneck to force the model to learn high-level video representations.
Note that these two decoder streams are only used during the training procedure. At test time, Rev2Net makes decisions without the pre-computed optical flow inputs. In other words, it avoids the disturbance of optical flow data, which has been shown in the cross-dataset experiment to harm the feature transferability.

Why is the multi-task framework useful to learn more generalizable features. Because it encourages the encoder stream to simultaneously learn short-term and long-term features from these two auxiliary tasks. We believe that comprehensively capturing the temporal relations at multiple time-scales can effectively cover various video dynamics across different video data sources. Further, the multi-task framework also enables knowledge transfer from the self-supervisions, from the short-term video dynamics from the optical flow data to the encoded frame features through an end-to-end training.
Note that the decoders are removed at test time, and thus Rev2Net can be seen as an I3D model at test time.

\vspace{-5pt}
\subsection{Decoding Discrepancy Penalty} 
\vspace{-5pt}

While traditional two-stream networks suffer from the decision discrepancy between the two network streams in the cross-dataset setting, Rev2Net avoids this issue by using only one frame encoder at test time. 
However, the self-supervised auxiliary tasks may introduce a new problem of decoding discrepancy to our proposed multi-task learning framework, i.e., the auxiliary tasks may not have a collaborative effect for learning features for the core classification task. 
In other words, though the flow prediction task and the reversed frame reconstruction task will force the encoder to focus on different parts of the input frames, they may not have the common positive and complementary effects on the training the encoder network.
To this end, we encourage the knowledge sharing between the two decoder streams in the training procedure, and penalize the distance of their features. We define a new training objective called the Decoding Discrepancy Penalty (DDP). We propose two forms of DDP respectively to the low-level and high-level features as shown in Figure \ref{F3D}.

\vspace{-5pt}
\vspace{10pt} \noindent \textbf{DDP Constraints on Low-Level Features.} We build the two decoder streams based on the low-level feature maps of the encoder (see Figure \ref{F3D}). In other words, both the decoders and the encoder have only a few convolutional layers. 
We use the Frobenius norm to penalize the distance of the decoders in the low-level feature space:
\begin{equation}
\label{equ:loss}
  \begin{aligned}
  & {\mathcal{L}}^{\textbf{low-level}}_\text{DDP} = \sum\limits_{{l} \in {\mathcal{L}}}{\|f_{\text{D}1}^l-f_{\text{D}2}^l\|_F}, \\
  \end{aligned} 
\end{equation}
where $\|\cdot\|_F$ is the Frobenius norm, $L$ is a set of convolutional layers included in the DDP, and $f_{\text{D}1}^l$ and $f_{\text{D}2}^l$ are the low-level feature maps of the optical flow decoder stream and the reversed frames decoder stream at layer $l$. 
As shown in Figure \ref{F3D}. There are two TransposeConv3D layers in the flow decoder and three of them in the frame decoder plus a Conv3D layer for generating the background which is not explicitly shown. These decoders are plugged into the original I3D network, standing on the feature maps of the $\texttt{Conv3d\_2c}$ layer, taking a feature volume of $16\times56\times56$ as the inputs. Intuitively, we do not want the decoders to be too strong, since training a good feature encoder may require relatively weak decoders. 

\vspace{-5pt}
\vspace{10pt} \noindent \textbf{DDP Constraints on High-Level Features.} We can also mitigate the high-level feature learning discrepancy by allocating more layers into the encoder network and its reversed decoder counterparts. 
In this method, particularly, the encoder has three parts of outputs: the features that are the inputs of the classifier, the mean $\mu_1$ and the variance $\sigma_1$ of the Gaussian distribution $\mathcal{N}(\mu_1,\sigma_1)$ that are used for optical flow prediction, and the mean $\mu_2$ and the variance $\sigma_2$ of the Gaussian distribution $\mathcal{N}(\mu_2,\sigma_2)$ that are used for reversed frames reconstruction.
We propose to apply KL divergence on the high-level, low-dimensional outputs of the encoder, i.e., $\mu_1,\sigma_1,\mu_2,\sigma_2$, to close the distance of $\mathcal{N}(\mu_1,\sigma_1)$ and $\mathcal{N}(\mu_2,\sigma_2)$: 
\begin{equation}
\label{equ:loss}
  \begin{aligned}
  & {\mathcal{L}}^{\textbf{high-level}}_\text{DDP} = D_\text{KL}(\mathcal{N}(\mu_1,\sigma_1) \| \mathcal{N}(\mu_2,\sigma_2)). \\
  \end{aligned} 
\end{equation}

Along with the optical flow prediction objective function or the reversed frames reconstruction objective function, the decoders can be trained similarly as variational autoencoders. 
By applying DDP to the overall loss function in the training process, including both the Frobenius norm and the KL divergence, we allow the two decoder streams to negotiate and collaborate to improve the consistency of their effects on learning a better encoder.

\vspace{-5pt}
\vspace{10pt} \noindent \textbf{Final Objective.} The final Rev2Net architecture with Frobenius norm DDP and KL divergence is shown in Figure \ref{F3D}. 
The three tasks, i.e., action recognition, flow prediction, frame reconstruction; along with their corresponding five loss function terms, including the DDP losses, can be jointly trained as
\begin{equation}
\label{equ:loss}
  \begin{aligned}
  \mathcal{L}(\mathcal{I},\mathcal{O},y) & = \mathcal{L}_\text{CE}(c,y) + \alpha{\mathcal{L}}^{\textbf{high-level}}_\text{DDP} +  \beta{\mathcal{L}}^{\textbf{low-level}}_\text{DDP} \\
  & + \lambda_\text{flow}\|\mathcal{O}-\widehat{\mathcal{O}}\|_F + \lambda_\text{im}\|\mathcal{I}-\widehat{\mathcal{I}}\|_F, \\
  \end{aligned} 
\end{equation}
where $\mathcal{L}_\text{CE}$ is the cross-entropy loss between the softmax output of the classifier $c$ and the ground truth action label $y$.
$\widehat{\mathcal{O}}=\{{\widehat{\mathcal{O}}_1}, \ldots ,{\widehat{\mathcal{O}}_\tau}\}$ are the generated optical flow, and $\mathcal{O}=\{{\mathcal{O}_1}, \ldots ,{\mathcal{O}_\tau}\}$ are the target optical flow which are pre-computed using TLV1 \cite{zach2007duality}. $\lambda_\text{flow}$ is the loss weight for the optical flow prediction task.
Similarly, $\widehat{\mathcal{I}}=\{{\widehat{\mathcal{I}}_1}, \ldots ,{\widehat{\mathcal{I}}_\tau}\}$ are generated frames, and $\mathcal{I}=\{{\mathcal{I}_1}, \ldots ,{\mathcal{I}_\tau}\}$ are real input frames. $\lambda_\text{im}$ is the loss weight for frames reconstruction task.
The hyper-parameters are found using the coarse-to-fine grid-search approach. We first search them using a coarse grid of $\{0, 0.0001, 0.001, 0.1, 1, 10\}$, and then locate the exact values using a fine grid of $\{0, 0.25, \ldots, 1\}$. Finally, we obtain hyper-parameters of $\beta=0.1$, $\alpha=0.01$, $\lambda_\text{im}=0.1$, $\lambda_\text{flow}=0.1$.

\vspace{-5pt}
\section{Experiments}
\vspace{-5pt}
In this section, we first introduce the datasets and compare Rev2Net with state-of-the-art video classification models for classic action recognition. The compared models include TSN \cite{wang2016temporal}, ActionFlowNet \cite{ng2018actionflownet}, Sun-OFF \cite{Sun_2018_CVPR}, TSM \cite{lin2018temporal}, and STM \cite{jiang2019stm}. We also perform ablation studies to explore the respective effectiveness of the multi-task framework and the DDP training objective. Finally, we use the cross-dataset experiments between UCF101 and HMDB51 to validate the generalization ability of Rev2Net. 

\vspace{-5pt}
\subsection{Datasets}
\vspace{-5pt}
We use the following datasets for the classic action recognition, and use the last two of them for the cross-dataset experiments: 
\begin{itemize}
\vspace{-5pt}
  \item UCF101 \cite{soomro2012ucf101} that contains $13{,}320$ annotated YouTube video clips from $101$ categories. Each video clip lasts $3$--$10$ seconds and consists of $100$--$300$ frames.
  \vspace{-5pt}
  \item HMDB51 \cite{Kuehne12} that contains $6{,}766$ video clips and covers $51$ action categories. Note that the videos from these datasets are distinct in both scenes and action dynamics.
  \vspace{-18pt}
  \item Kinetics \cite{kay2017kinetics} that contains $306{,}245$ annotated video clips from $400$ action categories.
\end{itemize}
\vspace{-5pt}

\begin{table}[t]
  \centering
   \caption{Results on the UCF101 and HMDB51 datasets. All compared models are pretrained on Imagenet.
   }
  \vspace{5pt}
   \label{final-result}
  \begin{tabular}{lccc}
    \toprule
    Model & Input & UCF101 & HMDB51 \\
    \midrule
    TSN & RGB            & 86.4 & 53.7 \\
    ActionFlowNet  & RGB       & 83.9 & 56.4 \\
    I3D  & RGB          & 84.5 & 49.8 \\    
    Sun-OFF   & RGB        & 93.3 & - \\
    Rev2Net w/o frame dec. & RGB & 93.8 & 69.6 \\
    Rev2Net w/o flow dec.  & RGB & 93.1 & 67.5 \\
    Rev2Net w/o DDP  & RGB & 93.3 & 69.7 \\
    Rev2Net (ours) & RGB & \textbf{94.6} & \textbf{71.1} \\
    \midrule
    TSN  & R + F  & 94.0 & 69.4 \\
    I3D  & R + F  & 93.4 & 66.4 \\
    Sun-OFF  & R + F & 96.0 & 74.2 \\
    Rev2Net  & R + F &   \textbf{97.1} & \textbf{78.0} \\
    \bottomrule
  \end{tabular}
  \centering
  \vspace{-5pt}
\end{table}

\begin{table}[t]
  \centering
   \caption{
   Results on the UCF101 and HMDB51 datasets. All compared models are pretrained on Kinetics.
   }
   \vspace{5pt}
   \label{result-pretrain-kinetics}
  \begin{tabular}{lccc}
    \toprule
    Model & Input & UCF101 & HMDB51  \\
    \midrule
    TSM & RGB  & 94.5 & 70.7 \\
    STM & RGB  & 96.2 & 72.2 \\
    I3D & RGB          & 95.6 & 74.8 \\
    Rev2Net  & RGB & \textbf{97.7} & \textbf{78.3} \\
    \midrule
    I3D & RGB + Flow & 98.0 & 80.7 \\
    Rev2Net  & RGB + Flow &  \textbf{98.7} & \textbf{81.9} \\
    \bottomrule
  \end{tabular}
  \centering
  \vspace{-5pt}
\end{table}

\begin{table}[t]
  \centering
   \caption{
   Results on the Kinetics dataset. All compared models are pretrained on ImageNet.}
   \label{kinetics-result}
   \vspace{5pt}
  \begin{tabular}{lcc}
    \toprule
    Model & Input & Kinetics\\
    \midrule
    I3D  & RGB           & 71.1\\
    TSM  & RGB & 72.5 \\ 
    Rev2Net w/o DDP & RGB           & 72.7 \\
    Rev2Net & RGB  &  \textbf{73.3} \\
    \midrule
    I3D   & RGB + Flow     & 74.2\\
    Rev2Net w/o DDP & RGB + Flow  & 74.1 \\
    Rev2Net & RGB + Flow    & \textbf{74.8}  \\
    \bottomrule
  \end{tabular}
  \centering
  \vspace{-10pt}
\end{table}

\begin{table}[t]
  \centering
  \caption{Cross-dataset results. We apply DANN \cite{ganin2015unsupervised} to all compared models to further enhance the transferability.}
  \vspace{5pt}
   \label{transfer-2}
  \begin{tabular}{lccc}
    \toprule
    Model & Input & HMDB $\rightarrow$ UCF & UCF $\rightarrow$ HMDB \\
    \midrule
    TSN  & RGB         & 58.5 & 40.1  \\
    TSM & RGB         & 58.9 & 41.2  \\
    I3D  & RGB         & 56.4 & 41.0  \\   
    \midrule
    TSN & R + F & 60.9 & 40.3  \\
    TSM & R + F & 61.2 &  41.5 \\
    I3D & R + F  & 57.9 & 41.5  \\ 
    \midrule
    Rev2Net & RGB  & \textbf{63.1} & \textbf{46.8}  \\ 
    \midrule
  \end{tabular}
 \vspace{-10pt}
\end{table}

\vspace{-5pt}
\subsection{Classic Action Recognition Results}
\vspace{-5pt}

\noindent \textbf{Comparing with Existing Models.}
Above all, we show that Rev2Net outperforms the state-of-the-art action recognition deep networks on each of the UCF101 and HMDB51 datasets, with either ImageNet or Kinetics pretrained models (Table \ref{final-result} and Table \ref{result-pretrain-kinetics}). 
It also achieves competitive results on the large scale Kinetics dataset, as shown in Table \ref{kinetics-result}. Rev2Net consistently performs better than other models with the same input modalities, including its base model, I3D.
Furthermore, we observe that introducing another I3D stream with optical flow inputs brings additional benefit to Rev2Net. 

\vspace{5pt} \noindent \textbf{Ablation Studies.} We evaluate the Rev2Net models that are trained in different self-supervised, multi-task manners: with or without the frame/flow decoder, and with or without the DDP loss.
Table \ref{final-result} suggests that, first, the frame reconstruction task and the flow prediction task makes individual contributions to the final result, and are complementary to each other. Removing either of the frame decoder or the flow decoder at training time will degenerate the performance of the Rev2Net.
Also, a Rev2Net model that is trained without the DDP loss performs even worse than the Rev2Net model that is trained only with the frame decoder ($93.3\%$ vs. $93.8\%$ on UCF101), indicating that the multi-task framework may have negative effect without the DDP constraint. 
In contrast, the DDP strategy could boost the performance of the final Rev2Net which is a multi-task framework ($93.3\%$ vs. $94.6\%$ on UCF101).

\vspace{-5pt}
\subsection{Cross-Dataset Action Recognition Results}
\vspace{-5pt}

In Section \ref{sec:cross}, we have found that the two-stream networks cannot perform well in the cross-dataset experiment setting because the optical flow features are difficult to be transferred. Rev2Net model resolves this problem by enhancing the generalization ability of features from the frame stream, and supervising the network with the optical flows instead of taking them as inputs. 
From Table \ref{transfer-2}, Rev2Net remarkably outperforms the TSN, TSM, and I3D models, indicating a strong generalization ability. It improves the base model, I3D, by $\textbf{11.9\%}$ upon its classification accuracy from UCF101 to HMDB51, and $\textbf{14.1\%}$ vice versa, with the same inputs of RGB frames.

\vspace{-5pt}
\section{Conclusions}
\vspace{-5pt}
In this paper, we designed a cross-dataset experiment to evaluate the generalization ability of video recognition models, and observed that the features extracted from the optical flow data are less generalizable.
To address the above problems and increase the generalization ability of an action recognition model, we proposed Rev2Net, a new multi-task learning framework for action recognition. 
Further, we proposed the decoding discrepancy penalty to encourage the collaboration of the training procedures of multiple self-supervised tasks.
In the experiments, we showed that our Rev2Net model trained with the DDP constraint outperforms the state-of-the-art methods on three datasets: UCF101, HMDB51, and Kinetics. Our model specifically shows a strong generalization ability in the setting of cross-dataset video action recognition.

\vspace{-5pt}
\section{Acknowledge}
\vspace{-5pt}
This work was supported in part by the Natural Science Foundation of China (61772299, 71690231), and the MOE Strategic Research Project on Artificial Intelligence Algorithms for Big Data Analysis.

\bibliographystyle{IEEEbib}
\bibliography{icme2020template}

\end{document}